\documentclass[]{spectral_ai}

\usepackage[toc,page,header]{appendix}

\usepackage{wrapfig}
\usepackage{multirow}
\usepackage{makecell}
\usepackage{colortbl}
\usepackage{amsfonts}
\usepackage{amsmath}
\usepackage{float}

\definecolor{colorfirst}{rgb}{.866,.945, 0.831} % green
\definecolor{colorsecond}{rgb}{1, 0.98, 0.83} % yellow
\definecolor{colorthird}{rgb}{0.76, 0.87, 0.92} % blue
\definecolor{colorcite}{rgb}{0.212, 0.490, 0.741}

\usepackage[most]{tcolorbox}
\usepackage{listings}

\newtcolorbox{promptbox}[2][]{
  breakable,
  colback=gray!5,
  colframe=gray!50,
  title=#2,
  fonttitle=\bfseries,
  #1
}

\usepackage{listings}
\usepackage{xcolor}

\lstdefinelanguage{json}{
    basicstyle=\ttfamily\small,
    showstringspaces=false,
    breaklines=true,
    frame=single,
    columns=fullflexible,
    morestring=[b]",
    stringstyle=\color{black},
    literate=
     *{0}{{0}}{1}
      {1}{{1}}{1}
      {2}{{2}}{1}
      {3}{{3}}{1}
      {4}{{4}}{1}
      {5}{{5}}{1}
      {6}{{6}}{1}
      {7}{{7}}{1}
      {8}{{8}}{1}
      {9}{{9}}{1}
      {:}{{:}}{1}
      {,}{{,}}{1}
      {\{}{{\{}}{1}
      {\}}{{\}}}{1}
}

\title{Towards Generalizable and Evidential Nuclear Magnetic Resonance-Based Molecular Structure Elucidation via Large Language Model Agent}
\author[1]{Zheng Fang}
\author[1]{Chen Yang}
\author[1]{Yusen Tan}
\author[4]{Yunpeng Zhao}
\author[3]{Fanjie Xu}
\author[7]{Hongxin Xiang}
\author[5]{Hanyu Sun}
\author[6]{Hanyu Gao}
\author[5]{Xiaojian Wang}
\author[2]{Wenjie Du}
\author[8]{Yuqiang Li}
\author[1, 6]{Jun Xia}
\affiliation[1]{Information Hub, The Hong Kong University of Science and Technology (Guangzhou), Guangzhou, China}

\affiliation[2]{State Key Laboratory of Precision and Intelligent Chemistry, University of Science and Technology of China, Hefei, China} 

\affiliation[3]{State Key Laboratory of Physical Chemistry of Solid Surface, College of Chemistry and Chemical Engineering, Xiamen University, Xiamen, China} 

\affiliation[4]{The University of Hong Kong, Hong Kong, China} 

\affiliation[5]{Peking Union Medical College and Chinese Academy of Medical Sciences, Beijing, China} 

\affiliation[6]{The Hong Kong University of Science and Technology, Hong Kong, China} 

\affiliation[7]{Hunan University, Hunan, China} 

\affiliation[8]{AI for Science Center, Shanghai Artificial Intelligence Laboratory, Shanghai, China} 

\abstract{
Nuclear Magnetic Resonance (NMR) spectroscopy is the gold standard for molecular structure elucidation, yet interpreting complex spectra for unknown molecules remains a bottleneck reliant on human expertise. While artificial intelligence has advanced this field, current methods face a critical trade-off: database retrieval cannot identify novel scaffolds, while \textit{de novo} molecular structure elucidation models operate as black boxes, lacking the atom-level interpretability required for rigorous scientific validation. Here, we present NMRAgent, an evidential reasoning agent powered by large language models (LLMs) that bridges this gap by integrating specialized spectral analysis tools with chemical knowledge graphs. Unlike previous approaches, NMRAgent mimics the deductive reasoning of human experts: it takes experimental NMR spectra and molecular formula as input, plans the elucidation process, proposes candidate structures, verifies peak-atom consistency, and refines misaligned substructure through formula-aware fragment optimization. Enabled by its evidential reasoning, NMRAgent outperforms state-of-the-art methods, improving top-1 accuracy by 46.5\% and Tanimoto similarity by 0.502 on a scaffold-split benchmark with novel scaffolds in the test set. Besides, we demonstrate the agent’s practical utility by elucidating the structures of two previously unknown natural products isolated from \textit{Hydrangea davidii} and \textit{Vitex trifolia}, and by correcting structural misassignments in established literature. By combining high-accuracy prediction with transparent and evidence-based reasoning, NMRAgent establishes a new paradigm for interpretable AI in analytical chemistry.
}

\correspondence{
Jun Xia (\url{junxia@hkust-gz.edu.cn})}


\begin{document}
% Custom commands for SpectralAI Template

% Theorem environments
\newtheorem{example}{Example}
\newtheorem{definition}{Definition}
\newtheorem{lemma}{Lemma}
\newtheorem{theorem}{Theorem}
\newtheorem{proposition}{Proposition}
\newtheorem{corollary}{Corollary}[proposition]
\newtheorem{assumption}{Assumption}
\newtheorem{observation}{Observation}

% Reference shortcuts
\newcommand{\fig}[1]{Fig.~\ref{#1}}
\newcommand{\eq}[1]{Eq.~(\ref{#1})}
\newcommand{\tb}[1]{Tab.~\ref{#1}}
\newcommand{\se}[1]{Section~\ref{#1}}
\newcommand{\ap}[1]{Appendix~\ref{#1}}

% Math shortcuts
\newcommand*{\dif}{\mathop{}\!\mathrm{d}}
\newcommand{\bbE}{\ensuremath{\mathbb{E}}}
\newcommand{\bbR}{\ensuremath{\mathbb{R}}}
\newcommand{\caL}{\ensuremath{\mathcal{L}}}
\newcommand{\caD}{\ensuremath{\mathcal{D}}}

% Comment command (for review process)
% \newcommand{\todo}[1]{\textcolor{red}{[TODO: #1]}}

\maketitle

% Optional: Add a teaser figure
% \begin{figure}[h]
%     \centering
%     \includegraphics[width=\linewidth]{figure/your_teaser.pdf}
%     \caption{Teaser figure description.}
%     \label{fig:teaser}
% \end{figure}

\section{Introduction}
Nuclear Magnetic Resonance (NMR) spectroscopy is a cornerstone of molecule structure elucidation, providing rich information about molecular connectivity, stereochemistry, and local chemical environments \cite{emwas2020nmr}. 
Among NMR techniques, $^{1}$H and $^{13}$C NMR spectra are particularly widely used in routine organic and natural-product analysis because of their accessibility and diagnostic value \cite{jonas2022prediction}.
Traditional NMR structure elucidation combines empirical spectral interpretation, quantum-chemical prediction, and database-assisted matching. While chemists use chemical-shift trends, multiplicity patterns, and coupling information to infer structural fragments \cite{ning2011interpretation,field2012organic}, Density Functional Theory (DFT)-based NMR calculations provide accurate candidate validation and stereochemical ranking \cite{ditchfield1974self,wolinski1990efficient}, and databases such as nmrshiftdb enable spectrum--structure dereplication \cite{steinbeck2003nmrshiftdb}. Nevertheless, these approaches remain constrained by expert labor, computational cost, and limited database coverage, motivating scalable data-driven methods for automated NMR-based structure elucidation.

\begin{figure}[htbp]
    \centering
    \includegraphics[width=\linewidth]{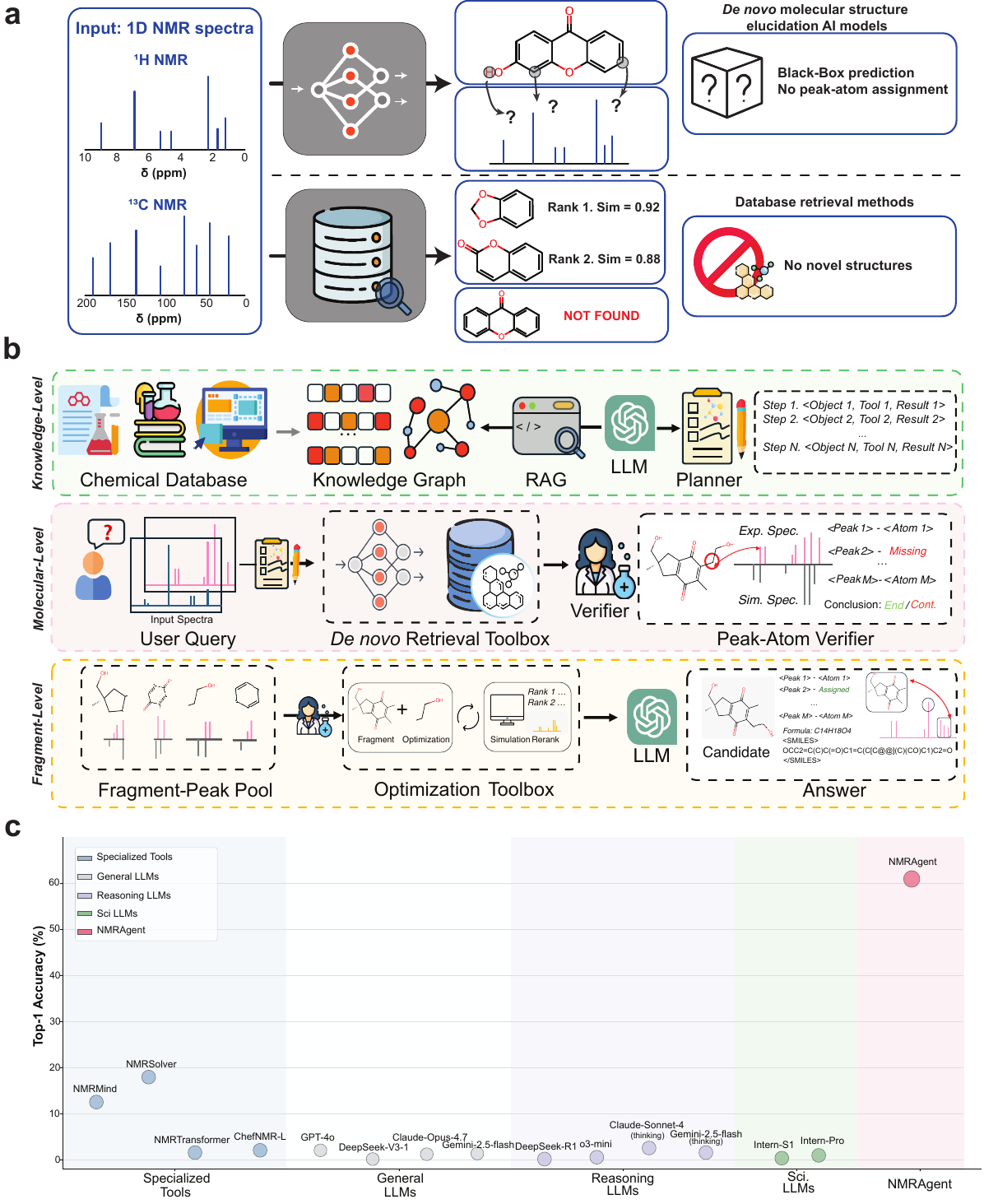}
    \vspace{-3mm}
    \caption{
    \textbf{Overview of previous methods and our proposed NMRAgent.}
    \textbf{a.} Limitations of previous NMR structure elucidation methods.
    \textbf{b.} The overall workflow of NMRAgent. 
    (1) {Knowledge-Level}: A Planner leverages RAG over chemical knowledge graph to formulate step-by-step elucidation plans. 
    (2) {Molecular-Level}: A diverse \textit{de novo} and retrieval toolbox generates candidates which are then rigorously validated by a {peak-atom verifier} through experimental-to-simulated spectral matching. 
    (3) {Fragment-Level}: For unresolved structures, the agent autonomously invokes an optimization toolbox to perform fragment optimization, guided by missing peak evidence.
    \textbf{c.} {NMRAgent achieves state-of-the-art performance on real-world datasets.} 
    Comparison of Top-1 Accuracy (\%) across five categories: Specialized Tools, General LLMs, Reasoning LLMs, Science LLMs, and our proposed NMRAgent. 
    While existing methods—including specialized computational tools and LLMs—struggle to exceed a 20\% accuracy threshold, NMRAgent significantly outperforms all baselines, achieving a breakthrough SOTA performance of over 60\% in real-world scenarios. 
  }
    \label{fig1}
\end{figure}

% Motivated by the transformative impact of deep learning in life sciences~\cite{jumper2021highly}, 
% [Polish note]
% This paragraph is overloaded and currently behaves like a related-work mini
% survey rather than an intro positioning paragraph.
% Problem:
%   too many model names and architectural details are packed into one long
%   paragraph, which weakens the transition into the paper's actual gap.
% Suggested rewrite direction:
%   synthesize prior work into 3 buckets only:
%   (1) sequence/generative models,
%   (2) diffusion or structure-generation models,
%   (3) retrieval/search-based systems.
% Mention representative citations, but move most model-by-model detail to
% Related Work.
% Suggested paragraph skeleton:
%   "Recent learning-based methods for NMR structure elucidation fall into
%   generative and retrieval-oriented paradigms. Generative models translate
%   spectra into molecular representations directly, whereas retrieval systems
%   search large compound-spectrum databases and optionally refine the returned
%   candidates. These methods have improved automation, but they still leave a
%   substantial gap between prediction and chemically justified elucidation."
% Recent research in NMR analysis has increasingly focused on the pivotal challenge of \textbf{{structure elucidation}}. The primary objective is the inverse mapping from experimental NMR spectra to molecular structures. 
To address these limitations, recent studies have increasingly explored \textit{de novo} molecular structure elucidation models based on deep learning, reframing automated NMR-based structure elucidation as an inverse mapping from NMR spectra to molecular structures.
A common line of work formulates this inverse problem as sequence-to-sequence generation, where NMR spectra are encoded into neural representations and molecular structures are decoded as SMILES strings~\cite{weininger1988smiles}.
For instance, CLAMS~\cite{tan2025clams} processes NMR spectra as 2D images using CNNs, whereas NMRTransformer~\cite{nmrformer} employs a 1D-CNN architecture to automatically learn spectral tokenization. Models such as NMR2Struct~\cite{hu2024accurate} and NMRMind~\cite{xue2025nmrmind} provide more explicit representations by encoding specific chemical shift ranges and intensities for transformer attention. Recently, inspired by breakthroughs in generative modeling~\cite{hoogeboom2022equivariant, morehead2024geometry}, diffusion-based paradigms have emerged. DiffNMR~\cite{yang2025diffnmr} progressively denoises and generates molecular 2D graphs, while ChefNMR~\cite{xiong2025atomic} generates 3D atomic conformations. 
Concurrently, retrieval-based methods such as NMRSolver~\cite{jin2025nmr} improve chemical validity by searching large-scale simulated spectral databases generated with fast machine-learning NMR predictors~\cite{xu2025toward, kim2023deepsat}, followed by fragment-based optimization.

Despite these rapid advancements, existing deep learning-based NMR structure elucidation methods still face two critical challenges that hinder their practical deployment. 
First, retrieval-based methods often achieve higher accuracy than \textit{de novo} models by searching over chemically valid reference databases. However, their performance is limited by the coverage and quality of these databases. 
When the target molecule contains a rare scaffold or represents a newly isolated natural product outside the retrieval space, these methods cannot easily propose genuinely novel structures.
Second, most \textit{de novo} molecular structure elucidation models operate largely as black-box systems. They directly output molecular structures while bypassing the peak-to-atom assignment step that is central to conventional NMR structure verification, limiting the interpretability of how the observed spectral signals support the predicted structure.
To address these challenges, we propose \textbf{NMRAgent}, a large-language-model-powered agentic framework designed for NMR structure elucidation. 
By leveraging the reasoning capability of large language model (LLM) together with specialized NMR analysis tools, NMRAgent introduces an {evidential reasoning} paradigm that explicitly identifies peak-atom assignments for structure verification. 
Unlike existing black-box models, this design grounds structural hypotheses in interpretable spectral evidence and supports more chemically reliable automated elucidation. More specifically, we design a three-level hierarchical architecture. At the \textbf{\textit{Knowledge-Level}}, a LLM planner utilizes Retrieval-Augmented Generation (RAG) over external chemical knowledge graph to formulate logical step-by-step elucidation plans. 
%
% Guided by these plans, the \textbf{\textit{Molecular-Level}} constructs a broad yet chemically meaningful candidate space by combining \textit{de novo} generation with retrieval-based tools. 
% Retrieval provides database-grounded candidates, whereas \textit{de novo} generation allows NMRAgent to explore structures beyond direct database matches. 
% Because generated structures do not naturally contain peak-atom annotations, fast NMR prediction models are used to estimate atom-resolved chemical shifts for each candidate. 
% This enables both \textit{de novo} and retrieved candidates to be jointly re-ranked in a unified evidence space according to spectral consistency and molecular structural similarity. 
% The resulting candidate pool is then examined by the Peak-Atom Verifier, which converts candidate-level predictions into explicit peak-atom assignments for interpretable structural reasoning.%
Guided by these plans, the \textbf{\textit{Molecular-Level}} constructs a broad yet chemically plausible candidate space by combining \textit{de novo} generation with retrieval tools. 
Retrieval provides database-grounded candidates, while \textit{de novo} generation enables exploration beyond direct database matches.
Using fast NMR prediction models, generated structures are annotated with atom chemical shifts and then re-ranked with retrieved candidates using a learned spectral-structural representation that considers both NMR consistency and molecular similarity. 
The resulting candidate pool is then examined by an LLM-powered Peak-Atom Verifier, which produces explicit peak--atom assignments to support interpretable structural reasoning.
Finally, if structural discrepancies persist, the \textbf{\textit{Fragment-Level}} performs evidence-guided structural optimization. Leveraging the LLM's chemical reasoning capability, NMRAgent interprets verifier-provided peak-atom mismatches as refinement hypotheses and applies a formula-aware heuristic to select and recombine compatible fragments from candidate pools. 
This work brings a new perspective to NMR-based molecular structure elucidation by introducing the first agentic framework that coordinates chemical knowledge, specialized tools, and spectral evidence for interpretable structural reasoning, offering a foundation for future advances in automated spectroscopy-driven chemical analysis.
% Instead of relying on blind exhaustive enumeration, this level focuses the search on chemically plausible variants supported by both molecular formula constraints and local spectral evidence.
% In summary, the main contributions of this work are threefold:
% \begin{itemize}
%     \item We propose \textbf{NMRAgent}, a novel agentic framework for NMR structure elucidation that transforms traditional black-box generation into an {evidential interpretive reasoning} process, providing explicit peak-atom assignments for rigorous structure verification.
%      \item We design a {three-level hierarchical architecture} (Knowledge-, Molecular-, and Fragment-Level) that progressively refines structural candidates. Notably, at the Molecular-Level, we leverage several search and \textit{De novo} generation tools together with large-scale self-supervised NMR encoders to expand the semantic search space and resolve structural divergence, while the Fragment-Level utilizes targeted local optimization to bypass combinatorial explosion..
%     \item Extensive experiments demonstrate that our approach achieves state-of-the-art performance on the serval benchmarks, delivering highly accurate and chemically reliable structural predictions with mechanistic transparency.
% \end{itemize}s

\section{Results and Discussion}

\subsection{Overview of NMRAgent}

As shown in Figure~\ref{fig1}b, we developed NMRAgent as an agentic framework for NMR-based molecular structure elucidation, designed to bridge candidate generation and evidence-based structural verification. 
Rather than treating elucidation as a single-step prediction problem, NMRAgent decomposes the task into a hierarchical workflow that mirrors expert chemical reasoning: retrieving relevant prior knowledge, proposing candidate structures, verifying atom-level spectral consistency, and refining unresolved regions.

% At the \textit{Knowledge-Level}, NMRAgent uses a retrieval-augmented planner to introduce external chemical knowledge into the elucidation process. 
% The planner queries chemical knowledge graph to retrieve relevant information, such as natural product classes, characteristic substructures, functional-group patterns, and reaction-related constraints. 
% The LLM then organizes this retrieved knowledge into a reasoning plan, which specifies which tools should be invoked and what structural evidence should be examined. 
% This level provides high-level chemical context and helps prevent the subsequent search from becoming a purely unconstrained enumeration problem.
At the {Knowledge-Level}, NMRAgent uses a retrieval-augmented planner to introduce external chemical knowledge into the elucidation process. 
% The planner queries a chemical knowledge graph to retrieve relevant contextual information, including compound-level annotations, scaffold-related records, functional-group clues, and source metadata. 
% This retrieval step supplements the LLM with domain-specific chemical context and helps mitigate its limited internal knowledge of fine-grained NMR spectroscopic evidence. 
%
The planner queries a chemical knowledge graph to retrieve relevant chemical context, including compound-level annotations, scaffold-related records, functional-group clues. 
This retrieval step supplements the LLM with domain-specific chemical context that are complementary to and coupled with NMR spectral evidence.
The LLM then organizes the retrieved information into a reasoning plan, specifying which tools should be invoked and what structural evidence should be examined. 
This level provides high-level chemical context and constrains subsequent molecular search beyond unconstrained enumeration.

% At the \textit{Molecular-Level}, NMRAgent constructs candidates from the input spectra. 
% We expand the Pubchem-NMRNet \cite{jin2025nmr} database to 158M pairs and pre-train a self-supervised NMR encoder on this database, which is then adapted to support both retrieval and fine-tuned to \textit{De novo} generation. 
% For retrieval, NMRAgent adopts a coarse-to-fine strategy: it first performs rapid search using Gaussian-based global NMR vectors to identify spectra-level neighbors, and then re-ranks these candidates using dense molecular embeddings to prioritize structures with closer chemical semantics. 
% In parallel, the encoder-conditioned \textit{De novo} generator directly proposes additional molecular structures from the input spectra. 
% All candidates are subsequently examined by the Peak-Atom Verifier through atom-annotated spectral simulation and peak-to-atom matching.
At the {Molecular-Level}, NMRAgent constructs a broad candidate pool from the input spectra by integrating retrieval and \textit{de novo} generation. 
We expand PubChem-NMRNet~\cite{jin2025nmr} to 158M spectrum--structure pairs and use deep-learning-based dense representations~\cite{ultranmr} to enable scalable molecular search.
To move beyond conventional spectrum-only retrieval, NMRAgent first retrieves spectra-level neighbors using Gaussian-based global NMR vectors and then re-ranks the corresponding structures with molecular dense embeddings, prioritizing candidates that are both spectrally consistent and chemically similar. 
Meanwhile, the \textit{de novo} generator proposes additional structures beyond database entries. 
Fast machine-learning NMR predictors~\cite{xu2025toward} are then used to produce atom-annotated spectra for retrieved and generated candidates, which are further examined by the LLM-powered Peak-Atom Verifier to establish explicit peak-atom correspondences.
% At the \textit{Molecular-Level}, NMRAgent executes the plan by constructing an initial pool of full-molecule candidates from the input spectra. 
% This level combines two complementary sources of candidates: retrieval-based search, which provides database-grounded chemically valid structures, and \textit{De novo} generation, which proposes structures beyond direct database matches. 
% To reduce retrieval-induced structural divergence, NMRAgent further employs a coarse-to-fine retrieval strategy, first identifying spectra-level neighbors and then prioritizing candidates with closer molecular semantics. 
% The resulting candidates are passed to the Peak-Atom Verifier, which simulates atom-annotated NMR shifts and compares them with the experimental peaks. 
% Rather than relying only on a global similarity score, the verifier produces explicit peak-to-atom assignments, indicating which parts of a candidate are supported by the spectrum and which local regions remain inconsistent.

At the {Fragment-Level}, NMRAgent uses the verifier output to perform targeted structural refinement when full-molecule candidates are not sufficiently supported. 
The candidate molecules are cleaved into chemically meaningful fragments, and the LLM interprets unresolved peak-atom mismatches to decide which fragments should be preserved and which regions should be exposed to repair. 
The optimization process is constrained by the target molecular formula and BRICS assembly rules, allowing bounded multi-fragment recombination while retaining spectrally supported substructures. 
Newly assembled candidates are then returned to the Peak-Atom Verifier, forming an iterative loop between atom-level spectral evidence and fragment-level optimization.

% This multi-level design provides two complementary advantages. 
% First, it expands the molecular search space by combining database-grounded retrieval, neural generation, and fragment-level optimization. 
% Second, it introduces interpretable spectral evidence into the reasoning loop through atom-level peak verification, allowing the system to assess when a candidate is sufficiently supported or when further refinement is needed. 
% Consistent with this design, NMRAgent substantially outperforms existing baselines in real-world structure elucidation settings. 
% As summarized in Figure~\ref{fig1}b, specialized tools and LLM-based methods generally achieve limited Top-1 accuracy, whereas NMRAgent reaches above 60\%, highlighting the effectiveness of integrating molecular search with evidence-guided reasoning and optimization.

\subsection{Generalization of NMRAgent Across Benchmarks}

\begin{figure}[]
    \centering
    \includegraphics[width=\linewidth]{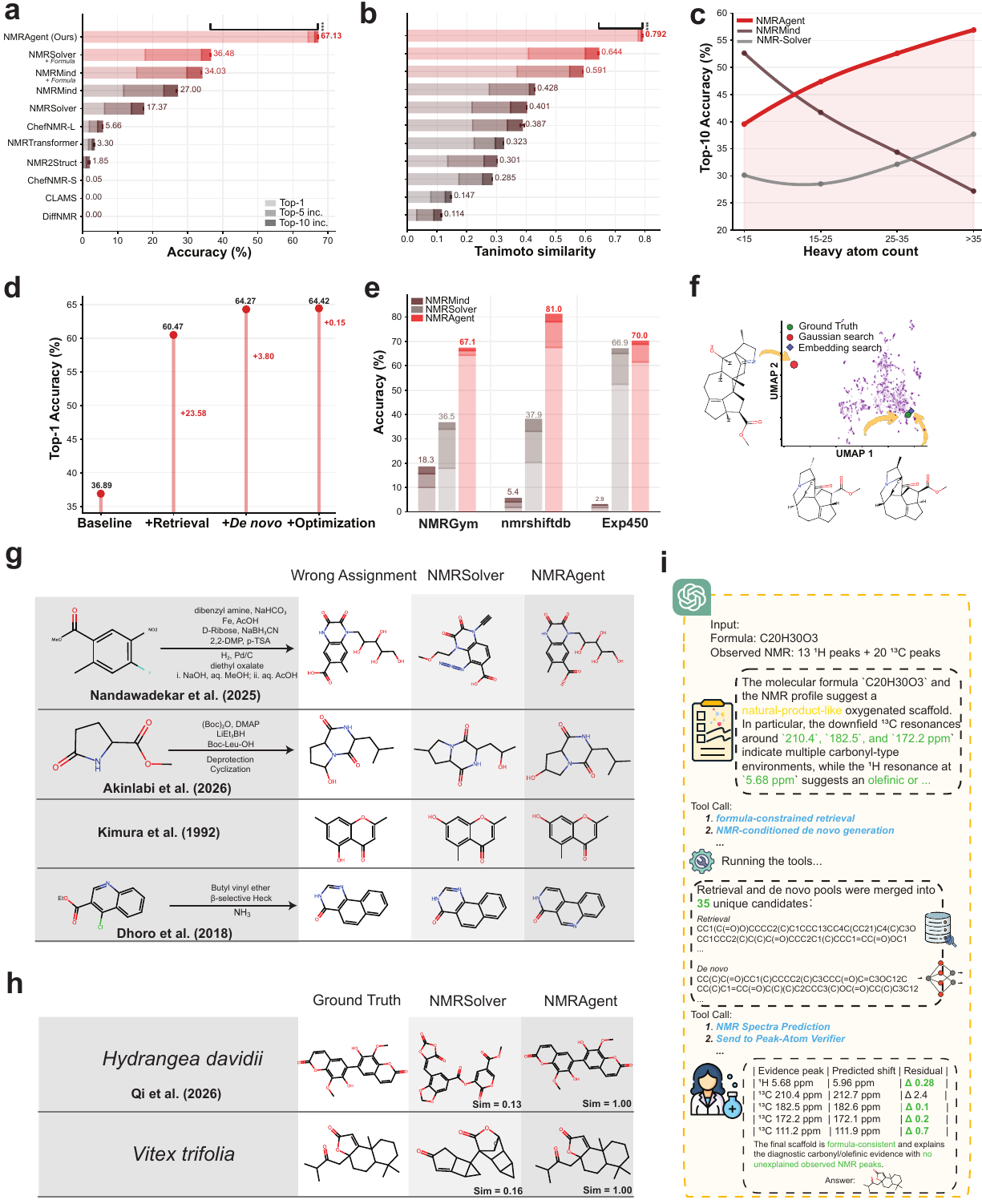}
    \vspace{-3mm}
    \caption{
        \textbf{Benchmark and real-world evaluation of NMRAgent.}
        \textbf{a.} Top-$k$ accuracy on NMRGym, showing that NMRAgent outperforms previous methods.
        \textbf{b.} Tanimoto similarity between predicted and ground-truth molecules on NMRGym, indicating improved structural proximity of NMRAgent predictions.
        \textbf{c.} Accuracy stratified by molecular size. Existing methods exhibit degraded performance as heavy atom count increases, whereas NMRAgent maintains stable accuracy on larger and more complex molecules.
        \textbf{d.} Ablation study of retrieval, \textit{De novo} generation, and fragment-level optimization, demonstrating the complementary contribution of each module.
        \textbf{e.} Evaluation across NMRGym, nmrshiftdb, and Exp450 benchmarks, highlighting the robustness of NMRAgent across diverse test sets.
        \textbf{f.} Embedding-space analysis showing that dense embedding retrieves structures closer to the ground truth.
        \textbf{g.} Representative example where NMRAgent corrects a literature-reported human peak-assignment error.
        \textbf{h.} Real-world natural product elucidation examples validated using literature reports and spectra provided by collaborating institution.
        \textbf{i.} An evidential reasoning trajectory of a real-world case.
        }
    \label{fig:2}
\end{figure}

\subsubsection{Evaluation on Scaffold-Split Benchmark}
We fine-tuned NMRAgent on the training split of NMRGym and evaluated its structure elucidation performance on the held-out scaffold-split test set. 
NMRGym is a benchmark for NMR-based molecular structure elucidation designed to assess generalization to unseen molecular scaffolds \cite{fang2026nmrgym}. 
Unlike random-split datasets, NMRGym separates molecules by scaffold, requiring models to infer structures that are topologically distinct from those seen during training or retrieval construction. 
This design better reflects real-world natural product elucidation, where target molecules often contain rare or previously unseen scaffolds rather than close database analogues. 
Additional details of NMRGym, including the fine-tuning setup, scaffold-split protocol, and molecular complexity statistics, are provided in the Supplementary Information.

We first assessed whether existing large language models can directly solve this task. 
As shown in Figure~\ref{fig1}c, general-purpose, reasoning-oriented, and science-oriented LLMs exhibit near-zero Top-1 accuracy in this setting. 
Although these models can occasionally succeed on simple cases and often produce fluent chemical explanations, they generally lack the spectral grounding required to recover exact molecular structures from NMR peaks. 
This result highlights a clear gap between language-level chemical reasoning and executable NMR structure elucidation: successful elucidation requires not only chemical knowledge, but also specialized molecular search, spectral simulation, and quantitative evidence verification.

We then compared NMRAgent with representative task-specific baselines on NMRGym, including retrieval-based methods and DL-based \textit{de novo} methods. 
As shown in Figure~\ref{fig:2}a, \textit{De novo} methods show limited accuracy under this scaffold-split setting, suggesting that direct spectrum-to-structure generation remains challenging for complex molecules with unseen scaffolds. 
Retrieval-based methods perform better because they search over chemically valid structures, but their performance remains constrained by database coverage and retrieval-induced structural divergence. 
In contrast, NMRAgent achieves a Top-10 accuracy of 67.13\%, substantially outperforming the strongest baseline, NMRSolver+Formula, which reaches 36.48\%. 
This corresponds to an absolute improvement of 30.65 percentage points and a relative improvement of approximately 84.0\%. 
The same trend is observed in structural similarity evaluation: NMRAgent achieves the highest Tanimoto similarity to the ground-truth molecules, indicating that its predictions are not only more often exact, but also closer to the correct structures in chemical space when exact recovery is not achieved (Figure~\ref{fig:2}b).

We further analyzed model performance as molecular size increases. 
A major challenge in NMR elucidation is that larger molecules usually contain more complex scaffolds, denser peak patterns, and more ambiguous local chemical environments. 
As shown in Figure~\ref{fig:2}c, baseline methods exhibit a clear degradation in accuracy as the heavy atom count increases. 
This trend indicates that existing methods struggle to maintain reliable structure recovery when the molecular search space becomes larger and the spectra become more difficult to disambiguate. 
In contrast, NMRAgent maintains stable performance across molecular-size groups, suggesting that the integration of retrieval, \textit{de novo} generation, atom-level verification, and fragment-level optimization provides a more robust strategy for complex natural-product-like molecules.

To further understand the source of NMRAgent's improvement, we conducted ablation studies over the major components of the framework. 
As shown in Figure~\ref{fig:2}d, retrieval, \textit{de novo} generation, and fragment-level optimization contribute complementary gains. 
Retrieval provides database-grounded chemically valid candidates, \textit{de novo} generation expands the search beyond direct database matches, and fragment-level optimization repairs unresolved regions using molecular formula constraints and verifier-provided mismatch evidence. 
The progressive improvement across these settings indicates that NMRAgent benefits from an integrated elucidation workflow rather than a single dominant module.
The ablation also shows that the gain is not merely a consequence of enlarging the retrieval database. 
When the NMRSolver-style retrieval baseline is evaluated with the expanded database, its Top-1 accuracy already exceeds the Top-10 accuracy obtained using the original database, confirming the importance of database coverage. 
Nevertheless, NMRAgent further improves over this expanded-database baseline, suggesting that the proposed representation, verification, and optimization mechanisms provide additional benefits beyond candidate pool expansion.
We further provide a qualitative case study to analyze the retrieval behavior. 
As shown in Figure~\ref{fig:2}f, dense embedding search retrieves a candidate that is structurally closer to the ground-truth molecule. 
This observation suggests that the learned embedding space better preserves molecular semantic similarity while maintaining spectral consistency, thereby reducing retrieval-induced structural divergence.
Additional case study visualizations are provided in the Supplementary Information.

\subsubsection{Evaluation on Additional Benchmarks}
We further assessed NMRAgent on nmrshiftdb and Exp450 to evaluate its robustness across different data sources. 
nmrshiftdb is a commonly used public NMR database \cite{steinbeck2003nmrshiftdb}, whereas Exp450 is a literature-derived experimental benchmark \cite{jin2025nmr}. 
Compared with NMRGym, these datasets present different challenges: nmrshiftdb contains many entries with incomplete spectral information, often including only a single spectrum type, while Exp450 reflects curated experimental cases collected from published reports.

As shown in Figure~\ref{fig:2}e, NMRAgent achieves the strongest performance on both datasets. 
On nmrshiftdb, NMRAgent reaches a Top-10 accuracy of 81.0\%, compared with 37.9\% for NMRSolver and 5.4\% for NMRMind. 
The large improvement indicates that NMRAgent can effectively handle heterogeneous public-database spectra, even when complete paired $^1$H and $^{13}$C information is not always available. 
This robustness is consistent with the design of the dense NMR encoder, which learns continuous spectral representations rather than relying solely on fixed sparse peak-vector matching.
On Exp450, NMRAgent obtains a Top-10 accuracy of 70.0\%, outperforming NMRSolver and NMRMind. 
Notably, NMRSolver remains competitive on Exp450, reflecting the advantage of retrieval-based methods in literature-derived benchmarks. 
However, NMRAgent further improves over this strong retrieval baseline by integrating coarse-to-fine retrieval, \textit{de novo} candidate generation, and optimization. 
These results show that NMRAgent is not limited to a single benchmark distribution, but can maintain strong performance across public database settings and experimentally curated literature cases.

\subsection{Real-world Application of NMRAgent}

\noindent\textbf{NMR Assignment Correction and Structural Revision.} 
Incorrect NMR-based structural assignments remain a persistent challenge in natural product chemistry, especially when regioisomers, scaffold isomers, or closely related analogues exhibit highly similar one-dimensional NMR spectra. 
To evaluate whether NMRAgent can assist structural revision, we therefore re-analyzed four representative structural revision cases, including Altechromone A~\cite{kimura1992altechromones,konigs2010structural}, Samoquasine A~\cite{morita2000samoquasine,timmons2008density,dhoro2018confirmation}, a natural tetrahydroquinoxaline-6-carboxylic acid from {Caulis Sinomenii}~\cite{nandawadekar2025structural}, and C5-hydroxy-cyclo(L-Pro-L-Leu)~\cite{akinlabi2026synthesis}.
These cases cover common sources of NMR misassignment, including positional ambiguity, regioisomeric uncertainty, and insufficient discrimination between structurally close candidates.
For each case, NMRAgent was provided only with the reported $^1$H and $^{13}$C NMR chemical shifts, without access to the subsequent synthetic, crystallographic, or DFT evidence used in the human revisions. 
Despite this restricted input, NMRAgent recovered the revised structures rather than the originally proposed assignments. 
These results suggest that NMRAgent can serve as a practical tool for detecting potentially incorrect NMR assignments and prioritizing chemically plausible structural revisions.

\noindent\textbf{Novel Natural Product Structure Elucidation.} 
Beyond correcting previously misassigned structures, we further evaluated NMRAgent in novel natural product elucidation scenarios. 
We considered two representative novel-structure cases: one coumarin dimer isolated from \textit{Hydrangea davidii}~\cite{qi2026coumarin}, and one compound isolated from \textit{Vitex trifolia}.
These cases represent more realistic application settings, where the target molecules are not simple benchmark compounds but structurally complex natural products requiring interpretation from experimental NMR evidence.
NMRAgent successfully recovered structures consistent with the literature-reported assignment and the collaborating group's experimentally validated solution, respectively. 
These results demonstrate that NMRAgent can extend beyond retrospective benchmark evaluation and support practical structure elucidation of newly isolated natural products. We show that NMRAgent demonstrates evidential reasoning capability, explicitly tracing how observed NMR signals support the predicted structure, as illustrated in Figure ~\ref{fig:2}i.

In addition to structure-level elucidation, we also explored whether the learned NMR representation could provide tentative metadata annotations for newly isolated compounds. 
Specifically, we integrated the knowledge-graph database described above with a fine-tuned NMR encoder, using approximately 1,000 labeled examples per class to train lightweight classifiers for chemical superclass, natural product pathway, and genus-level metadata. 
Rather than serving as definitive annotations, these predictions are intended to provide auxiliary hypotheses that may help contextualize novel molecules in broader natural-product chemical space. 
The resulting embedding-space visualizations, including t-SNE plots, are provided in the Supplementary Information.

\section{Discussion and Limitations}
\label{sec:conclusion}
Despite its strong performance, NMRAgent still has several limitations. 
First, the current framework focuses on one-dimensional $^1$H and $^{13}$C NMR spectra, while expert elucidation often relies on richer spectroscopic evidence, including 2D NMR, IR, UV, and stereochemical measurements. 
Future versions should therefore incorporate additional tools and modalities to support more comprehensive spectral reasoning. 
Second, although the expanded database substantially improves candidate coverage, NMRAgent remains partially constrained by database completeness and still does not fully reach the reliability of experienced human experts, especially for rare scaffolds, noisy spectra, or highly ambiguous natural products. 
Third, the LLM component does not yet intrinsically understand NMR spectral information; its current effectiveness mainly comes from tool grounding, retrieval, and verifier-provided evidence. 
Future work should explore agentic learning and post-training strategies that allow LLMs to learn from verifier feedback, failed reasoning trajectories, and expert corrections, thereby internalizing spectroscopic knowledge more effectively.
Despite these limitations, NMRAgent establishes an LLM-powered agentic framework for evidence-grounded NMR structure elucidation. 
By moving beyond direct black-box prediction toward interpretable candidate verification, our work provides a scalable foundation for incorporating richer spectral modalities, broader chemical knowledge, and stronger spectroscopic reasoning in future systems.

\section{Method}

\subsection{Preliminaries}
In standard structure elucidation pipelines, the exact molecular formula $\mathcal{F}$ is typically determined prior to NMR analysis, often utilizing mass spectrometry (MS) \cite{ning2011interpretation}. Given $\mathcal{F}$ as a global structural constraint, the subsequent task is to deduce the precise atomic connectivity.
We formally define the input NMR information as a collection of peak sets. 
Specifically, let $\mathcal{S}_H = \{s^H_1, s^H_2, \dots, s^H_{N_H}\}$ and 
$\mathcal{S}_C = \{s^C_1, s^C_2, \dots, s^C_{N_C}\}$ denote the $^1$H and $^{13}$C 
NMR spectra, respectively, where each peak $s_i \in \mathbb{R}$ represents 
a chemical shift value (in ppm). 
Detailed chemical definitions are provided 
in Appendix~\ref{chemical}. 
Since the physical measurement of NMR peaks is invariant to permutation, we 
treat each spectrum strictly as a set rather than a sequence. For 
methodological simplicity in the following sections, we denote the unified 
input as $\mathcal{S} = \mathcal{S}_H \cup \mathcal{S}_C$, assuming that 
both spectra are concurrently provided to the model. 
The corresponding molecular structure is represented as a SMILES \cite{weininger1988smiles} 
sequence $Y = \{y_1, \dots, y_L\}$, where each element $y_i \in \mathcal{V}$ denotes 
a discrete token from a predefined chemical vocabulary $\mathcal{V}$ (e.g., atom 
symbols, bonds, and structural indicators).

\subsection{Overall Workflow}
\label{sec:workflow}

To elucidate a molecular structure from NMR spectra, human chemists typically follow a deductive workflow that starts from global constraints such as the molecular formula, interprets local chemical environments from NMR peaks, and iteratively verifies structural hypotheses against experimental evidence \cite{ning2011interpretation,field2012organic,socha2023nmr}. 
Inspired by this process, NMRAgent adopts a three-level hierarchical workflow consisting of Knowledge-Level planning, Molecular-Level candidate generation and verification, and Fragment-Level optimization. 
An LLM acts as the reasoning controller throughout the workflow, coordinating knowledge retrieval, tool invocation, candidate evaluation, and evidence-based refinement.

As illustrated in Figure~\ref{fig1}, the \textbf{\textit{Knowledge-Level}} uses a retrieval-augmented planner to formulate a target-specific elucidation plan from external chemical knowledge. 
The \textbf{\textit{Molecular-Level}} constructs full-molecule candidates conditioned on the input spectra $\mathcal{S}$ and molecular formula $\mathcal{F}$, using a \textit{de novo} and  retrieval toolbox together with an LLM-powered {Peak-Atom Verifier} that establishes explicit peak-atom correspondences. 
If unresolved spectral evidence remains, the \textbf{\textit{Fragment-Level}} performs targeted fragment optimization guided by the missing or inconsistent peak evidence identified by the verifier. 
Detailed descriptions of each module are provided below.
% \begin{figure}
%     \centering
%     \includegraphics[width=\linewidth]{embedding.pdf}
%     \caption{Caption}
%     \label{fig:placeholder}
% \end{figure}

\subsection{Knowledge-Level Planner}
\label{knowledge}

In chemical research, expert chemists rely on standardized analytical workflows for molecular structure elucidation, grounded in established spectroscopic heuristics and empirical rules \cite{ning2011interpretation, field2012organic, socha2023nmr}. 
To emulate this human expertise, we propose the \textbf{Knowledge-Level Planner}, a central reasoning module within NMRAgent that uses Retrieval-Augmented Generation (RAG) to formalize complex structure elucidation processes. 
To establish a comprehensive knowledge base, we integrate heterogeneous scientific resources, including natural product repositories \cite{sorokina2021coconut, poynton2025natural, rutz2022lotus} and bioactivity-related databases \cite{gaulton2012chembl, chandak2023building}. 
In addition to external chemical knowledge, the planner maintains an NMR evidence memory that stores previously verified spectroscopic reasoning cases, such as confirmed peak--atom assignments, solvent conditions, multiplicity patterns, confidence scores, and expert or user feedback. 
This memory allows the planner to reuse reliable assignment evidence from prior elucidation cases, thereby making the reasoning process closer to the way chemists draw on accumulated spectroscopic experience.

Formally, the planning process is triggered by the molecular formula $\mathcal{F}$, input spectra $\mathcal{S}$, and experimental metadata $\mathcal{M}$, where $\mathcal{M}$ may include available contextual information such as reaction precursors, reagents, catalysts, biological source, or isolation conditions. 
We denote the external chemical knowledge graph as $\mathcal{K}$ and the NMR evidence memory as $\mathcal{H}$. 
Given $(\mathcal{F}, \mathcal{S}, \mathcal{M})$, the planner retrieves relevant chemical context from $\mathcal{K}$ and assignment-level evidence from $\mathcal{H}$:
\begin{equation}
\mathcal{C}_{K} = \mathcal{R}_{K}(\mathcal{K}; \mathcal{F}, \mathcal{M}), 
\qquad
\mathcal{C}_{H} = \mathcal{R}_{H}(\mathcal{H}; \mathcal{S}, \mathcal{F}, \mathcal{M}),
\end{equation}
where $\mathcal{C}_{K}$ denotes retrieved chemical knowledge and $\mathcal{C}_{H}$ denotes retrieved memory evidence from previous NMR assignment cases. 
The LLM then synthesizes these retrieved contexts into a procedural guideline $\mathcal{G}$:
\begin{equation}
\mathcal{G} = \mathrm{LLM}(\mathcal{F}, \mathcal{S}, \mathcal{M}, \mathcal{C}_{K}, \mathcal{C}_{H}).
\end{equation}

To translate the high-level guideline $\mathcal{G}$ into an executable workflow, the planner integrates it with a predefined toolset $\mathcal{T}$ and corresponding operation descriptions $\mathcal{A}$. 
It generates a molecule-specific elucidation plan 
$\mathcal{P} = \{ \mathcal{P}_1, \mathcal{P}_2, \dots, \mathcal{P}_n \}$, where each reasoning step is represented as:
\begin{equation}
\mathcal{P}_i = \langle t_i, x_i, e_i, y_i \rangle, 
\qquad t_i \in \mathcal{T},
\end{equation}
where $t_i$ is the selected tool, $x_i$ is the input to the tool, $e_i$ is the supporting evidence retrieved from $\mathcal{C}_{K}$ or $\mathcal{C}_{H}$, and $y_i$ is the expected structural inference or intermediate result. 
In this framework, the inclusion of experimental metadata and NMR evidence memory helps NMRAgent avoid blind structural enumeration. 
Instead, the planner produces a chemically grounded and evidence-supported elucidation plan that reflects both external chemical knowledge and reusable spectroscopic assignment experience.

\subsection{Molecular-Level Candidate Construction and Verification}

Recent \textit{de novo} models directly generate molecular strings or graphs while bypassing the peak-to-atom assignment process that is central to conventional NMR verification. 
Second, retrieval-based methods provide chemically valid candidates, but their performance depends on the coverage of the reference spectrum--structure database and on the quality of the retrieval representation. 
In particular, conventional spectrum-only retrieval can identify candidates with similar global peak distributions, but such candidates are not necessarily structurally close to the target molecule, especially for complex natural products with rare scaffolds or subtle local variations.

To address these issues, NMRAgent constructs a hybrid candidate pool by integrating database-grounded retrieval with \textit{de novo} generation. 
We expand Pubchem-NMRNet~\cite{jin2025nmr} to 158M spectrum--structure pairs by augmenting the original database with additional compounds, including complex scaffold-containing and natural-product-like molecules. 
For the expanded portion of the database, NMR spectra are simulated \textit{in silico} using Gaussian software~\cite{gaussian16}, enriching the reference space beyond the original database coverage. 
This expanded database provides a broader pool of chemically valid candidates for downstream retrieval and reasoning. 
In parallel, the \textit{de novo} and retrieval toolbox can invoke NMR-conditioned molecular generators, such as ChefNMR~\cite{xiong2025atomic} and NMRMind~\cite{xue2025nmrmind}, to propose additional structures beyond database entries. 
The retrieved and generated candidates are then merged into a unified candidate pool for verification.

To support more informative retrieval, we use UltraNMR~\cite{ultranmr}, a self-supervised representation model for NMR spectra. 
Given heterogeneous $^1$H and $^{13}$C NMR inputs, UltraNMR maps spectra into a dense continuous embedding space that captures both individual peak patterns and contextual dependencies across spectra. 
The model is pre-trained with a masked spectral modeling objective, where a subset of chemical-shift tokens is masked and predicted from the remaining spectral context. 
In addition, UltraNMR is trained to estimate structural similarity between molecules from their spectral embeddings:
\begin{equation}
    \mathcal{L}_{\mathrm{SSL}}
    =
    \mathcal{L}_{\mathrm{mask}}
    +
    \lambda_{\mathrm{FP}}\mathcal{L}_{\mathrm{FP}},
\end{equation}
where $\mathcal{L}_{\mathrm{mask}}$ denotes the masked chemical-shift prediction loss and $\mathcal{L}_{\mathrm{FP}}$ denotes the fingerprint-similarity prediction loss. 
This objective encourages the spectral representation to preserve structural information useful for candidate ranking.

To further distinguish structurally similar candidates, especially formula-identical isomers, we further incorporates a cross-modal contrastive objective:
\begin{equation}
    \mathcal{L}_{\mathrm{NCE}}
    =
    - \log
    \frac{
    \exp(\cos(\bm{z}, \bm{v}^{+})/\tau)
    }{
    \exp(\cos(\bm{z}, \bm{v}^{+})/\tau)
    +
    \sum_{k=1}^{K}
    \exp(\cos(\bm{z}, \bm{v}^{-}_{k})/\tau)
    },
\end{equation}
where $\bm{z}$ is the projected NMR embedding, $\bm{v}^{+}$ is the fingerprint representation of the matched molecule, $\bm{v}^{-}_{k}$ denotes the $k$-th hard-negative fingerprint representation, and $\tau$ is the temperature hyperparameter. 
The final representation learning objective is:
\begin{equation}
    \mathcal{L}_{\mathrm{UltraNMR}}
    =
    \mathcal{L}_{\mathrm{SSL}}
    +
    \lambda_{\mathrm{NCE}}\mathcal{L}_{\mathrm{NCE}}.
\end{equation}

For database-grounded candidate search, NMRAgent performs coarse-to-fine retrieval. 
At the coarse stage, the query spectrum is encoded into a global NMR vector and used for high-recall retrieval of spectra-level neighbors following the global vector retrieval protocol~\cite{jin2025nmr}. 
This step rapidly narrows the 158M-scale database to a manageable candidate set with similar overall NMR patterns. 
At the fine stage, the retrieved candidates are re-ranked using UltraNMR dense embeddings, prioritizing structures whose learned spectral representations are closer to the query. 
Thus, the coarse stage emphasizes efficient spectrum-level recall, whereas the fine stage refines the ranking with a representation that better reflects structural similarity.

In parallel with retrieval, UltraNMR is paired with an autoregressive molecular decoder for \textit{de novo} candidate generation. 
Given a target SMILES sequence $\bm{Y}=\{y_1,y_2,\ldots,y_L\}$ and the projected NMR embedding $\bm{z}$, the decoder models:
\begin{equation}
    p_{\theta}(\bm{Y}\mid \bm{z})
    =
    \prod_{t=1}^{L}
    p_{\theta}(y_t \mid y_{<t}, \bm{z}),
\end{equation}
with the training objective:
\begin{equation}
    \mathcal{L}_{\mathrm{gen}}
    =
    -
    \sum_{t=1}^{L}
    \log p_{\theta}(y_t \mid y_{<t}, \bm{z}).
\end{equation}
This generation branch complements retrieval by proposing additional structures that may not appear in the reference database.

% After candidate construction, NMRAgent applies an LLM-powered Peak-Atom Verifier to convert candidate scoring into explicit atom-level evidence. 
% For each retrieved or generated molecule, a fast machine-learning NMR predictor, such as NMRNet~\cite{xu2025toward}, is used to simulate atom-annotated chemical shifts. 
% The verifier compares the predicted shifts with the query spectrum using a set-similarity metric~\cite{jin2025nmr} and establishes peak-to-atom correspondences whenever the predicted and observed signals are compatible. 
% The resulting evidence includes candidate-level consistency scores, assigned peak--atom pairs, and unresolved or mismatched spectral regions. 
% These outputs are passed to the LLM for evidence aggregation and candidate prioritization. 
% If the verifier finds sufficient spectral support, NMRAgent returns a ranked elucidation list; otherwise, the unresolved peak evidence is used to guide subsequent Fragment-Level optimization. 
After candidate construction, NMRAgent applies an LLM-powered peak-atom verifier to convert candidate scoring into explicit atom-level evidence. 
Given the unified candidate pool $\mathcal{C}_{\text{pool}}=\{c_1,\ldots,c_M\}$ and the query spectrum $\mathcal{S}$, the verifier first invokes a fast machine-learning NMR predictor~\cite{xu2025toward, detanet} to generate atom-annotated chemical shifts for each candidate. 
For a candidate molecule $c$, this process produces $\hat{\mathcal{S}}^{\text{atom}}(c)=\{(\hat{\delta},a,\alpha)\}$, where $\hat{\delta}$ is a predicted chemical shift, $a$ is the associated atom index, and $\alpha\in\{^1\mathrm{H},^{13}\mathrm{C}\}$ denotes the nucleus type.

The verifier compares $\hat{\mathcal{S}}^{\text{atom}}(c)$ with the observed spectrum $\mathcal{S}$ using a set-similarity metric~\cite{jin2025nmr}. 
It then identifies compatible peak--atom correspondences, producing an assignment set $\mathcal{A}(c)$, a consistency score $s(c)$, and unresolved mismatch regions $\mathcal{U}(c)$. 
Together, these outputs form an evidence object:
\begin{equation}
    \mathcal{E}(c)
    =
    \{c,\ s(c),\ \mathcal{A}(c),\ \mathcal{U}(c)\}.
\end{equation}
The peak-atom verifier aggregates the verifier evidence across $\mathcal{C}_{\text{pool}}$ to produce either a ranked elucidation list or targeted mismatch descriptions for subsequent Fragment-Level optimization.

\subsection{Fragment-Level Optimization}

When the peak-atom verifier indicates insufficient spectral support for the current molecular candidates, NMRAgent activates the \textit{Fragment-Level Optimization} module for targeted structural refinement. 
Rather than treating refinement as unconstrained molecular generation, this module uses verifier-provided evidence to localize uncertain structural regions and explores chemically valid fragment substitutions or recombinations under molecular-formula constraints. 
This design complements molecular-level candidate construction by preserving spectrally supported regions as structural anchors and focusing refinement on unresolved or mismatched regions.

Given the candidate pool evaluated by the peak-atom verifier, NMRAgent constructs a fragment pool $\mathcal{P}_{\mathrm{frag}}$ by decomposing relevant candidates according to chemically defined fragmentation rules, such as BRICS-based cleavage or other rule-based bond-disconnection strategies. 
Each fragment is associated with its sub-formula and local spectral evidence inherited from the verifier. 
Specifically, atom-level shift annotations indicate which local regions are supported by matched experimental peaks and which regions remain ambiguous or inconsistent. 
The LLM acts as an evidence interpreter rather than a direct structure generator: it uses verifier outputs to identify fragments that should be preserved, regions that require modification, and alternative fragment compositions that should be explored.

Formally, let $\mathcal{F}_{\mathrm{target}}$ denote the target molecular formula. 
Based on the peak--atom evidence, NMRAgent identifies a set of supported fragments $\mathcal{P}_{\mathrm{keep}}$ and a candidate repair space $\mathcal{P}_{\mathrm{edit}}$ associated with unresolved spectral evidence. 
Fragment-Level Optimization searches for chemically assembleable fragment combinations that satisfy the formula constraint and improve agreement with the experimental spectrum:
\begin{equation}
    S^{*}
    =
    \arg\min_{S \subseteq \mathcal{P}_{\mathrm{frag}}}
    d_{\mathrm{set}}
    \left(
        \hat{\mathcal{Y}}_{\mathrm{NMR}}(\mathrm{Assemble}(S)),
        \mathcal{Y}_{\mathrm{exp}}
    \right),
\end{equation}
subject to:
\begin{equation}
    \sum_{f \in S} \mathrm{Formula}(f)
    =
    \mathcal{F}_{\mathrm{target}},
    \quad
    \mathcal{P}_{\mathrm{keep}} \subseteq S,
    \quad
    S \setminus \mathcal{P}_{\mathrm{keep}} \subseteq \mathcal{P}_{\mathrm{edit}}.
\end{equation}
Here, $\mathrm{Assemble}(\cdot)$ denotes rule-guided chemical assembly, $\hat{\mathcal{Y}}_{\mathrm{NMR}}(\cdot)$ denotes the predicted NMR shifts of the assembled molecule, $\mathcal{Y}_{\mathrm{exp}}$ is the experimental spectrum, and $d_{\mathrm{set}}(\cdot,\cdot)$ measures spectral disagreement. 
The formula constraint acts as a hard filter, while the verifier-guided preservation and repair constraints focus the search on regions with insufficient peak--atom evidence. 
The resulting formula-compatible and chemically assembleable candidates are then passed back to the peak--atom verifier for iterative evidence-based assessment.

\section{Code and Data Availability}
The code is available at \url{https://github.com/Meaw0415/NMRAgent}.
The checkpoints of UltraNMR~\cite{ultranmr} are available at \url{https://huggingface.co/milesyc/ultranmr}. 
The released NMRMind~\cite{xue2025nmrmind} checkpoints used in this work can be accessed through the official NMRMind repository at \url{https://github.com/WJmodels/NMRMind}. 
The released ChefNMR~\cite{xiong2025atomic} checkpoints are available at \url{https://zenodo.org/records/17766755}.
The released NMRNet~\cite{xu2025toward} checkpoints for forward NMR prediction are available at \url{https://zenodo.org/records/19142375}. 
The source data of the Pubchem-NMRNet database~\cite{jin2025nmr} can be accessed through \url{https://huggingface.co/datasets/yqj01/SimNMR-PubChem}. 
The NMRGym~\cite{fang2026nmrgym} dataset can be accessed through \url{https://huggingface.co/datasets/meaw0415/NMRGym}. 
The processed nmrshiftdb data~\cite{steinbeck2003nmrshiftdb} can be obtained at \url{https://zenodo.org/records/19142375}. 
The Exp450 dataset can be obtained at \url{https://doi.org/10.5281/zenodo.16952024}. 
Detailed information on the datasets is provided in the Supplementary Information.

\noindent\textbf{Acknowledgments.} 

\bibliographystyle{plainnat}
\bibliography{ref}

\clearpage
\beginappendix

\renewcommand\thefigure{\Alph{figure}}
\renewcommand\thetable{\Alph{table}}
\renewcommand\thesection{\Alph{section}}
\setcounter{figure}{0}
\setcounter{table}{0}
\setcounter{section}{0}

\section{Chemical Preliminaries}
\label{chemical}
Formally, we model the spectrum as a continuous function $x(\delta): \mathbb{R} \to \mathbb{R}$ over the chemical shift domain $\delta$. Disregarding spin-spin interactions, the signal is a superposition of $N$ independent resonance peaks:

\begin{equation}
    x(\delta) = \sum_{n=1}^{N} I_n \cdot \mathcal{L}(\delta; \mu_n, \lambda_n) + \xi(\delta),
    \label{eq:ideal_nmr}
\end{equation}

where $I_n$ and $\mu_n$ denote the intensity and chemical shift of the $n$-th nucleus, and $\xi(\delta)$ represents additive Gaussian noise. The spectral lineshape $\mathcal{L}$ typically follows a Lorentzian distribution with half-width $\lambda_n$:

\begin{equation}
    \mathcal{L}(\delta; \mu_n, \lambda_n) = \frac{1}{\pi} \frac{\lambda_n}{(\delta - \mu_n)^2 + \lambda_n^2}.
    \label{eq:lorentzian}
\end{equation}

In experimental settings, scalar coupling ($J$-coupling) introduces spin-spin interactions between neighboring nuclei, splitting the resonance signal into multiplets. The model generalizes to a nested summation over $K_n$ sub-peaks:
\begin{equation}
    x(\delta) = \sum_{n=1}^{N} \sum_{k=1}^{K_n} I_{n,k} \cdot \mathcal{L}(\delta; \mu_{n,k}, \lambda_{n}) + \xi(\delta),
    \label{eq:coupling_nmr}
\end{equation}

where the relative positions $\mu_{n,k}$ and intensities $I_{n,k}$ are governed by the coupling constants ($J$-values) and molecular topology.

Crucially, the accessibility of these spectral parameters varies significantly across data sources. High-fidelity datasets typically provide comprehensive annotations, including precise chemical shifts ($\mu$), peak intensities ($I$), and coupling constants ($J$).

This appendix provides supplementary material that supports the main paper.

% \subsection{Implementation Details}

\section{Implementation Details}

\subsection{Benchmark Datasets and Evaluation Metrics}

PubChem-NMRNet~\cite{jin2025nmr} was used as the simulated NMR reference database for retrieval-based candidate search. 
The original database contains approximately 100 million molecule--spectrum paired records from PubChem, with $^1$H and $^{13}$C NMR spectra predicted by NMRNet. 
In this work, we further extended the reference database to 158 million molecule--spectrum pairs.
NMRGym~\cite{fang2026nmrgym} was used as a scaffold-split benchmark to evaluate the generalization ability of NMRAgent, particularly its ability to recover novel molecular structures beyond seen scaffolds. 
The nmrshiftdb benchmark was originally derived from nmrshiftdb~\cite{steinbeck2003nmrshiftdb} and followed the processed version released with NMRNet~\cite{xu2025toward}. 
Exp450 was used as an external experimental benchmark curated from the NMRSolver study~\cite{jin2025nmr}, containing 450 literature-extracted samples that reflect real-world scenarios.

We evaluated performance using top-$k$ exact-match accuracy and top-$k$ molecular similarity, with $k\in\{1,5,10\}$. 
Given the limited availability of standardized 2D NMR benchmarks and the insufficient stereochemical information provided by 1D NMR spectra alone, stereochemistry was removed for both exact-match accuracy and Tanimoto-similarity evaluation.
Top-$k$ accuracy was defined as the fraction of test cases for which the ground-truth structure appeared in the top-$k$ ranked predictions:
\begin{equation}
\mathrm{Acc@}k
=
\frac{1}{N}
\sum_{i=1}^{N}
\mathbb{I}
\left[
G_i \in \{\hat{M}_{i,1}, \ldots, \hat{M}_{i,k}\}
\right].
\end{equation}

Molecular similarity was measured using Tanimoto similarity between binary Morgan fingerprints with radius 2 and 2048 bits. 
For two molecules $A$ and $B$, let $\mathbf{f}(A)$ and $\mathbf{f}(B)$ denote their Morgan fingerprints. 
The Tanimoto similarity is defined as
\begin{equation}
\mathrm{Tan}(A,B)
=
\frac{
\mathbf{f}(A)^\top \mathbf{f}(B)
}{
\|\mathbf{f}(A)\|_1+\|\mathbf{f}(B)\|_1-\mathbf{f}(A)^\top \mathbf{f}(B)
}.
\end{equation}
For top-$k$ molecular similarity, we reported the highest Tanimoto similarity between the ground-truth molecule and any of the top-$k$ ranked predictions, averaged over all test cases.

\subsection{Agent System Prompts}

To make the agent behavior auditable and reproducible, NMRAgent separates the system prompts from the graph execution logic. 
The prompts define the role, constraints, evidence requirements, and output format of each agent, while deterministic Python tools execute retrieval, \textit{de novo} generation, reranking, optimization, and molecular editing. 
Below, we provide the system prompts.

\subsubsection{Planner System Prompt}

\begin{promptbox}{Planner system prompt}
You are an expert organic chemistry NMR structure-elucidation planner.

Your task is to design an executable plan for elucidating the structure of an unknown molecule from its molecular formula, $^1$H NMR spectrum, $^{13}$C NMR spectrum, optional experimental metadata, previous tool outputs, verifier feedback, and recalled confirmed memories.

You should not directly decide the final molecular structure. 
Instead, you should determine which tools should be invoked, what evidence should be examined, and whether retrieval, \textit{de novo} generation, verifier-guided optimization, or a larger candidate pool is needed.

\textbf{Core instructions:}
\begin{enumerate}
    \item Treat the molecular formula as a hard constraint. 
    Do not plan any candidate search or structural modification that violates the molecular formula.

    \item Use the $^1$H NMR spectrum to reason about proton environments, integration, multiplicity when available, and repeated or equivalent signals. 
    If the input contains expanded repeated shifts, do not collapse repeated peaks unless explicitly instructed.

    \item Use the $^{13}$C NMR spectrum as strong evidence for carbon environments, including carbonyl-like carbons, aromatic or alkene carbons, oxygen-bearing carbons, and saturated aliphatic carbons.

    \item Use optional experimental metadata only as contextual evidence. 
    Metadata such as reaction precursors, reagents, catalysts, biological source, or isolation conditions can guide hypotheses, but cannot override the molecular formula or spectral evidence.

    \item Use recalled memories only as analogical evidence from previously confirmed cases. 
    A memory cannot override the query spectra, molecular formula, or verifier evidence.

    \item Preserve both retrieval and \textit{de novo} candidates when both are useful. 
    Retrieval rank alone should not suppress plausible \textit{de novo} candidates before peak--atom verification.

    \item Prefer a larger or more diverse candidate pool when the spectra suggest compact natural-product-like scaffolds, fused rings, lactones, enones, unusual oxygenation patterns, or other rare scaffold types.

    \item If previous verifier feedback identifies unmatched peaks, inconsistent local assignments, or unresolved mismatch regions, plan targeted optimization rather than blind regeneration.

    \item Return JSON only. 
    Do not include free-form text outside the JSON object.
\end{enumerate}
\end{promptbox}

\noindent The Planner returns the following JSON schema:

\begin{lstlisting}[language=json]
{
  "analysis": "short evidence-grounded reasoning about formula and NMR signals",
  "use_retrieval": "<true_or_false>",
  "use_denovo": "<true_or_false>",
  "retrieval_top_k": "<integer>",
  "denovo_top_k": "<integer>",
  "save_pool_file": "<true_or_false>",
  "need_large_pool": "<true_or_false>",
  "need_opt_after_generation": "<true_or_false>",
  "notes_for_executor": "concrete execution notes for tool use"
}
\end{lstlisting}

\subsubsection{Executor System Prompt}

\begin{promptbox}{Executor system prompt}
You are the execution controller for NMRAgent.

Your role is to execute the Planner's JSON instructions using deterministic tools. 
You do not decide the final molecular structure. 
You must preserve candidate provenance and return structured outputs for downstream peak--atom verification.

\textbf{Execution instructions:}
\begin{enumerate}
    \item Execute retrieval, \textit{de novo} generation, pool merging, reranking, optimization, and molecular editing only when requested by the Planner or by verifier feedback.

    \item Preserve candidate provenance whenever possible, including SMILES string, source label, source rank, source score, molecular formula, spectrum metadata, and candidate-pool path.

    \item Keep retrieval-derived candidates and \textit{de novo} candidates visible for downstream verification. 
    Do not discard a candidate solely because it comes from a lower-volume source.

    \item Deduplicate candidates by canonical non-isomeric SMILES while retaining all available source provenance.

    \item Do not silently invoke retrieval or \textit{de novo} generation inside an optimization step. 
    Candidate construction, candidate merging, optimization, and verification must remain auditable as separate operations.

    \item Optimization tools should operate only on an existing candidate pool.

    \item Verifier-guided in-place molecular editing may be invoked only when the verifier localizes a high-confidence mismatch to a specific atom or small chemical environment.

    \item After any local edit, retain the unedited parent candidate for comparison.

    \item If a tool fails, returns invalid output, or produces an empty candidate pool, report the failure explicitly in the structured execution output.

    \item Return structured execution results, including generated files, candidate counts, source composition, tool status, and any warnings needed by the verifier.
\end{enumerate}

You must not provide the final molecular structure unless the Peak--Atom Verifier has supplied sufficient evidence.
\end{promptbox}

\subsubsection{Peak--Atom Verifier System Prompt}

\begin{promptbox}{Peak--Atom Verifier system prompt}
You are an expert NMR peak--atom assignment verifier.

Your task is to evaluate candidate molecules using formula consistency, forward-predicted NMR shifts, matched peaks, unmatched peaks, residual errors, atom-level assignment summaries, and candidate provenance. 
You should not judge candidates from SMILES strings alone. 
Your verdict must be grounded in explicit spectral evidence.

\textbf{Evidence to inspect:}
\begin{enumerate}
    \item Molecular formula consistency between the query and each candidate.

    \item Overall NMR similarity score from the reranking tool.

    \item Matched query peaks and their assigned candidate atoms.

    \item Unmatched query peaks, especially diagnostic peaks that indicate missing functional groups or unresolved local environments.

    \item Unused predicted peaks, especially predicted signals that have no reasonable support in the experimental spectrum.

    \item $^1$H and $^{13}$C residuals, including whether the largest errors occur in chemically diagnostic regions.

    \item Atom-level assignment summaries, including which parts of the molecule are well supported and which regions remain uncertain.

    \item Candidate provenance, including retrieval, \textit{de novo}, optimized, merged, edited, or seed sources.

    \item Previous verifier outputs or recalled confirmed memories, if available. 
    These may provide supporting context but cannot override the current query evidence.
\end{enumerate}

\textbf{Decision instructions:}
\begin{enumerate}
    \item Return \texttt{accept} only when both $^1$H and $^{13}$C evidence are coherent and no major diagnostic query peaks remain unexplained.

    \item A \textit{de novo} candidate may be accepted over retrieval candidates if its formula consistency, spectral similarity, and peak--atom assignments provide stronger evidence.

    \item Return \texttt{need\_opt} when the best candidate is globally reasonable but contains local mismatches that can be targeted by fragment optimization or in-place editing.

    \item Return \texttt{need\_bigger\_pool} when the candidate pool lacks sufficiently diverse or formula-compatible alternatives.

    \item Return \texttt{need\_retry} when tool outputs are invalid, incomplete, inconsistent, or insufficient for evidence-based verification.

    \item When optimization is needed, provide concrete mismatch descriptions, including the relevant unmatched peaks, poorly matched atoms, or local structural regions.
\end{enumerate}

Return JSON only using the required schema.
\end{promptbox}

\noindent The Peak--Atom Verifier returns the following JSON schema:

\begin{lstlisting}[language=json]
{
  "verdict": "<accept | need_opt | need_bigger_pool | need_retry>",
  "analysis": "evidence-grounded summary of the decision",
  "top_candidate": "<SMILES string or null>",
  "retry_recommendation": "<specific next action or null>"
}
\end{lstlisting}

\subsection{Planner Tools}
NMRAgent uses a lightweight Graph RAG tool to provide advisory chemical context for the Planner. 
We build a heterogeneous chemical knowledge graph from public natural-product databases, molecular databases, ontology resources, biomedical relation graphs, and synthetic chemistry knowledge cards. 
Source-specific parsers convert CSV, JSON, SQLite, OBO graph, and tabular triple files into a unified graph containing molecule or ontology nodes, typed relation edges, retrievable text documents, and aliases. 
The merged graph contains 4,275,055 nodes, 2,516,946 edges, 4,337,742 documents, and 10,432,630 aliases. 
For runtime use, the graph is indexed with SQLite/FTS to support fast metadata lookup and BM25-style document retrieval. 
The retrieved KG evidence is provided to the Planner only as background context and cannot override molecular formula constraints, NMR reranking evidence, or peak-assignment diagnostics.

Beyond the external knowledge graph, NMRAgent further includes a lightweight memory store for real-world iterative structure elucidation. The memory module is implemented with \texttt{langgraph.store.memory.InMemoryStore}, where confirmed cases are stored under \texttt{("nmr\_agent", "confirmed\_cases")}. Auto-remembering accepted outputs is disabled by default; instead, cases are written to memory only after explicit user confirmation, particularly when the final structure and peak--atom assignments have been checked. This prevents unverified agent outputs from becoming self-reinforcing evidence.
Each memory record stores the molecular formula, observed $^1$H and $^{13}$C shifts, final SMILES, optional canonical SMILES, verified peak--atom assignments, diagnostic notes, source information, and metadata. During inference, relevant memories are recalled using a local non-embedding similarity score that combines formula matching with greedy one-to-one matching of $^1$H and $^{13}$C peaks under ppm tolerances.
Recalled memories are provided to the Planner and Verifier as \texttt{relevant\_confirmed\_memories}. They serve only as advisory evidence and cannot override hard constraints such as molecular formula consistency, candidate validity, or NMR reranking results. This design allows NMRAgent to reuse user-validated experimental knowledge while reducing confirmation bias and error propagation.

\subsection{Retrieval Tool}

Following NMRSolver~\cite{jin2025nmr}, we implement a retrieval tool based on large-scale NMR spectral matching. 
For each modality $m\in\{H,C\}$, the input spectrum is represented as a peak set 
$\mathcal{S}_m=\{s_i^m\}_{i=1}^{N_m}$. 
We first convert the discrete peaks into a continuous signal by Gaussian convolution:
\begin{equation}
g_{\mathcal{S}_m}(t)
=
\sum_{i=1}^{N_m}
\exp\left(
-\frac{(t-s_i^m)^2}{2\sigma_m^2}
\right),
\end{equation}
where $\sigma_m$ controls the peak width. We set $\sigma_H=1.0$, and $\sigma_C=10.0$ in all NMR2Vector calculations.
The continuous signal is then discretized at $K=128$ uniformly sampled points to form the NMR2Vector representation:
\begin{equation}
\mathbf{v}_{\mathcal{S}_m}
=
[g_{\mathcal{S}_m}(t_1),g_{\mathcal{S}_m}(t_2),\ldots,g_{\mathcal{S}_m}(t_K)].
\end{equation}
Given a database spectrum $\mathcal{R}_m=\{r_j^m\}_{j=1}^{M_m}$, its vector $\mathbf{v}_{\mathcal{R}_m}$ is constructed in the same way. 
The vector similarity is computed as:
\begin{equation}
\mathrm{VecSim}(\mathcal{S}_m,\mathcal{R}_m)
=
\frac{
\mathbf{v}_{\mathcal{S}_m}^{\top}
\mathbf{v}_{\mathcal{R}_m}
}{
\|\mathbf{v}_{\mathcal{S}_m}\|_2
\|\mathbf{v}_{\mathcal{R}_m}\|_2
}.
\end{equation}
For peak-level comparison, we also use Set Similarity~\cite{jin2025nmr}. 
The similarity between $\mathcal{S}_m=\{s_i^m\}_{i=1}^{N_m}$ and $\mathcal{R}_m=\{r_j^m\}_{j=1}^{M_m}$ is defined as:
\begin{equation}
\mathrm{SetSim}(\mathcal{S}_m,\mathcal{R}_m)
=
\frac{1}{\sqrt{N_mM_m}}
\max_{P}
\sum_{(i,j)\in P}
f(s_i^m,r_j^m),
\end{equation}
where $P$ denotes a one-to-one matching between peaks, and
\begin{equation}
f(s_i^m,r_j^m)
=
\exp\left(
-\frac{(s_i^m-r_j^m)^2}{2\sigma_m^2}
\right).
\end{equation}
In addition to NMR2Vector, we use UltraNMR~\cite{ultranmr} as a dense NMR spectral encoder. 
Following the original UltraNMR setting, we pre-train the encoder on 1.58M NMR spectra and use it to encode all spectra in the reference database. 
The resulting dense embeddings are indexed with FAISS~\cite{douze2025faiss} for approximate nearest-neighbor search. 
During inference, NMRAgent follows a two-stage retrieval strategy. 
First, NMR2Vector is used to retrieve an initial pool of candidates by direct spectral similarity. 
Second, the pool is reranked using UltraNMR dense embeddings, which capture learned spectral-structural similarity and help prioritize candidates that are not only peak-wise similar but also chemically plausible. 
When the molecular formula $\mathcal{F}$ is available, we further apply a formula filter to prioritize formula-consistent candidates before downstream verification.

\subsection{\textit{De novo} Tools}
For \textit{de novo} candidate generation, NMRAgent integrates NMRMind~\cite{xue2025nmrmind}, ChefNMR~\cite{xiong2025atomic}, and UltraNMR~\cite{ultranmr}. 
We use the released checkpoints of NMRMind and ChefNMR directly, without additional training. UltraNMR is fine-tuned only on the NMRGym training split to avoid test-set leakage. 
Following the original setting, SMILES are tokenized with the ChemBERTa vocabulary~\cite{chithrananda2020chemberta}, and $\mathcal{F}$ is encoded as a 50-dimensional atom-count vector, projected by a two-layer MLP, and used as a prefix token for generation. 
We set learning rate $1\times10^{-5}$, batch size 64, 50 epochs, dropout 0.1, weight decay 0.01, a 6-layer Transformer decoder, and length penalty 1.0. We use beam search during inference, 
\subsection{Spectral Simulation Tools}
For forward NMR prediction, NMRAgent uses NMRNet~\cite{xu2025toward} as the spectral simulation tool. We directly use the released NMRNet checkpoint without additional training. 
\subsection{Optimization Tools}
In addition to fragment recombination, NMRAgent also implements a lightweight in-place editing mode for cases where the verifier localizes the mismatch to a specific atom or a small chemical environment, such as neighboring atoms, an adjacent substituent, or a local functional group. 
This mode is distinct from fragment-level recombination: fragment optimization explores alternative fragment compositions when the mismatch spans a broader local region, whereas in-place editing performs minimal local modifications when the evidence supports a more specific edit. 
This design is useful because rule-based fragmentation may be too coarse to express small corrections, such as replacing a single heteroatom or removing an unsupported atom. 
The LLM can invoke constrained RDKit-based operations, including SMILES canonicalization, atom replacement, and atom deletion. 
After each edit, the molecule is sanitized and canonicalized, while the original parent candidate is retained for comparison. 
Edited candidates are then returned to the peak-atom verifier for reassessment, ensuring that local modifications are accepted only when they improve the evidence-grounded spectral consistency.

\section{Experimental NMR Spectra of Two Novel Natural Products}

\textbf{Compound 1}

SMILES: \texttt{CC12C(CCC(O3)(CC(C(C)C)=O)C2=CC3=O)C(C)(C)CCC1}

Molecular formula: {C20H30O3}.

$^1$H NMR (400 MHz, CDCl$_3$) $\delta$ 5.68 (s, 1H), 3.19--2.98 (m, 2H), 2.68 (p, $J=6.9$ Hz, 1H), 2.59 (dt, $J=9.1, 2.4$ Hz, 1H), 1.86 (dt, $J=13.7, 4.1$ Hz, 2H), 1.77 (tt, $J=14.5, 3.9$ Hz, 1H), 1.69--1.57 (m, 2H), 1.56--1.44 (m, 3H), 1.32--1.23 (m, 1H), 1.20 (s, 3H), 1.11 (dd, $J=6.9, 2.9$ Hz, 6H), 1.02 (td, $J=8.1, 3.8$ Hz, 1H), 0.94 (d, $J=3.7$ Hz, 6H).

$^{13}$C NMR (101 MHz, CDCl$_3$) $\delta$ 210.38, 182.46, 172.23, 111.19, 86.84, 56.39, 46.03, 42.31, 41.72, 40.55, 40.20, 37.25, 34.24, 33.51, 21.71, 19.45, 18.34, 18.07, 17.90, 17.51.

\vspace{0.5em}

\textbf{Compound 2}

SMILES: \texttt{COc1c(O)c(-c2cc3ccc(=O)oc3c(OC)c2O)cc2ccc(=O)oc12}

Molecular formula: {C20H14O8}.

$^1$H NMR (400 MHz)$\delta$ 7.90 (d, $J=9.3$ Hz), 7.39 (s), 6.12 (d, $J=9.3$ Hz), 3.96 (s).

$^{13}$C NMR (101 MHz)$\delta$ 164.0, 110.4, 147.1, 112.0, 125.5, 129.0, 160.0, 137.8, 149.1, 164.0, 110.4, 147.1, 112.0, 125.5, 129.0, 160.0, 137.8, 149.1, 61.3, 61.3.

\section{More Results}
\begin{figure}
    \centering
    \includegraphics[width=\linewidth]{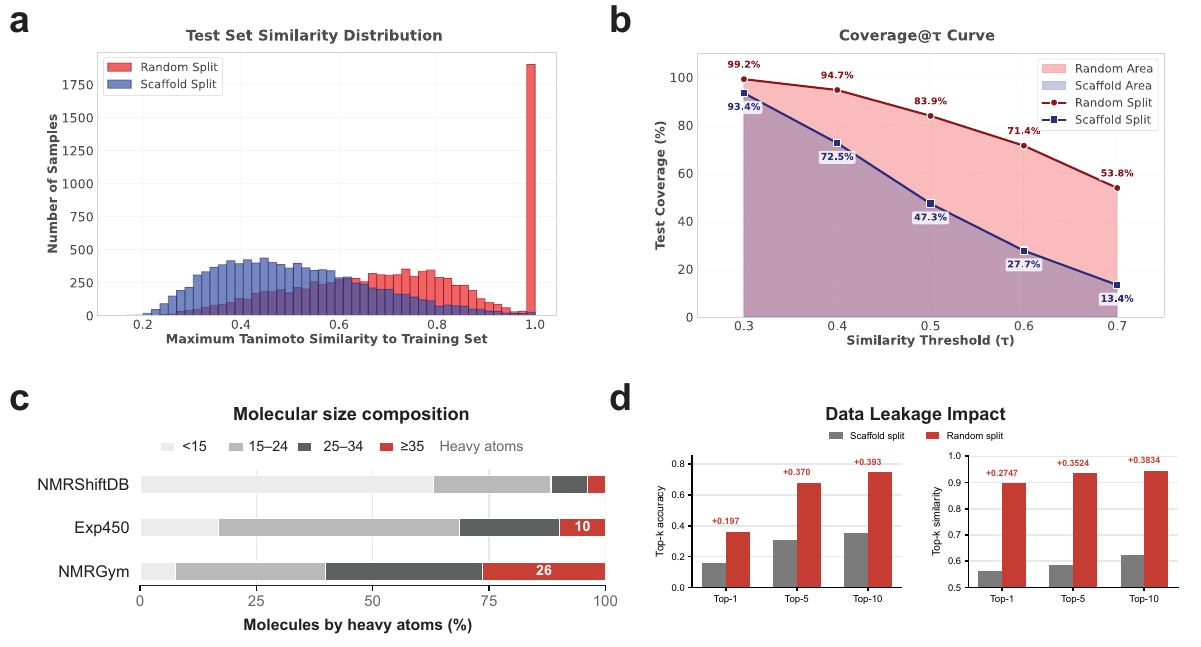}
    \caption{
        \textbf{Benchmark composition and scaffold-split evaluation analysis of NMRGym.}
        \textbf{a.} Distribution of the maximum Tanimoto similarity between each test molecule and the training set under random-split and scaffold-split settings. Compared with random splitting, the scaffold split substantially reduces near-duplicate test molecules and provides a more stringent evaluation of structural generalization.
        \textbf{b.} Coverage@$\tau$ curves under different similarity thresholds, showing that random splits retain much higher training-set coverage of test molecules, whereas scaffold splits better separate test structures from training scaffolds.
        \textbf{c.} Molecular size composition of nmrshiftdb, Exp450, and NMRGym, grouped by heavy atom count. NMRGym contains a larger fraction of complex molecules with high heavy-atom counts, making it a more challenging benchmark for natural-product-like structure elucidation.
        \textbf{d.} Impact of data leakage on model evaluation. Random splitting leads to substantially higher Top-$k$ accuracy and structural similarity than scaffold splitting, highlighting the importance of scaffold-aware evaluation for realistic NMR structure elucidation.
        }
    \label{fig:3}
\end{figure}

\begin{figure}
    \centering
    \includegraphics[width=\linewidth]{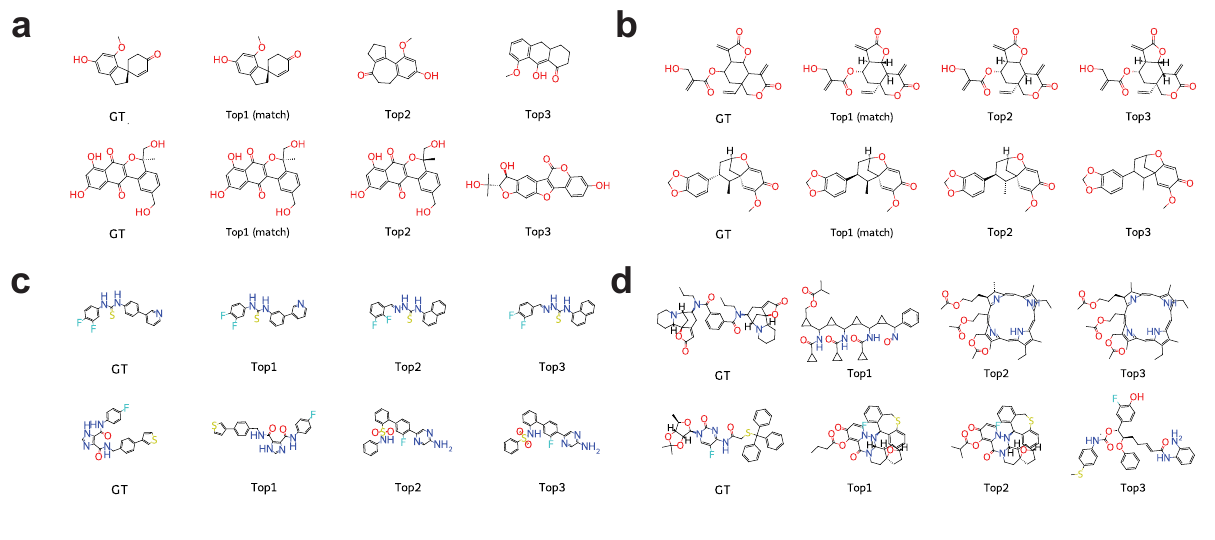}
    \caption{
         \textbf{Qualitative analysis of NMRAgent structure elucidation results.}
         \textbf{a.} Representative successful cases in which NMRAgent recovers the exact ground-truth structures, including stereochemical assignments, at Top-1.
        \textbf{b.} Cases in which NMRAgent recovers the correct molecular connectivity but fails to fully match the stereochemical configuration.
        \textbf{c.} High-similarity failure cases.
        \textbf{d.} Low-similarity failure cases.
        }
    \label{fig:3}
\end{figure}

\begin{figure}
    \centering
    \includegraphics[width=\linewidth]{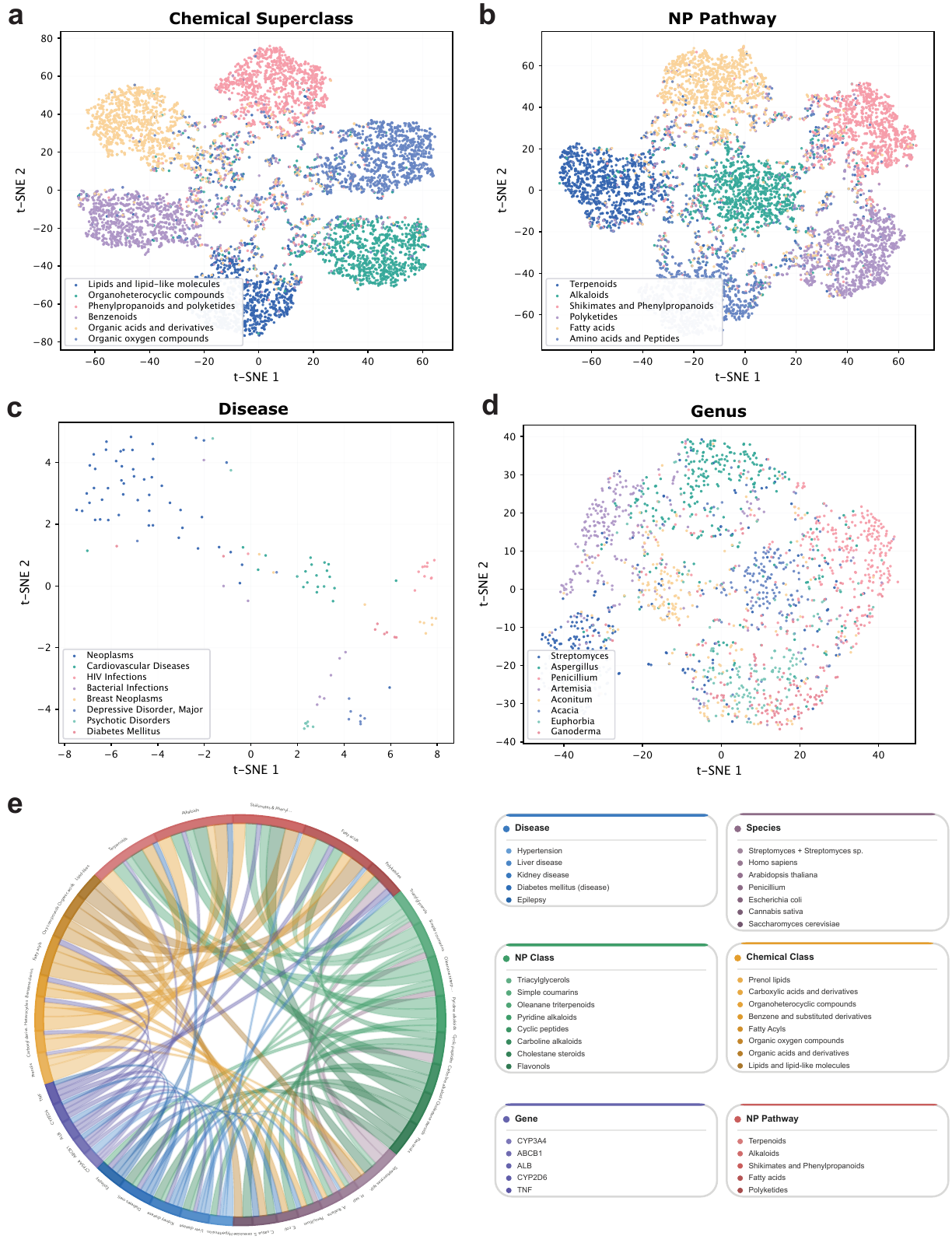}
    \caption{\textbf{Visualization of NMR embedding spaces and knowledge-graph-derived metadata annotations.} 
    \textbf{a-d.} The t-SNE plots show representative organization of molecules with respect to chemical superclass, natural product pathway, disease association, and genus-level metadata. 
    \textbf{e.} The chord diagram highlights the diverse metadata labels captured by our knowledge graph across biological, chemical, and natural-product-related categories.}
    \label{fig:placeholder}
\end{figure}

\clearpage
\newpage
\section{Full Numerical Results}
\vspace{-3mm}
\begin{table}[h]
\caption{Comparison of structural elucidation performance on the NMRGym dataset.}
\vspace{-3mm}
\centering
\label{nmr2smiles1}
\setlength{\tabcolsep}{8pt} 
\resizebox{0.8\linewidth}{!}{
\begin{tabular}{l c@{\hskip 0.2cm}c c@{\hskip 0.2cm}c c@{\hskip 0.2cm}c}
\toprule
\toprule
 & \multicolumn{2}{c}{Top-1} & \multicolumn{2}{c}{Top-5} & \multicolumn{2}{c}{Top-10} \\
\cmidrule(lr){2-3} \cmidrule(lr){4-5} \cmidrule(lr){6-7}
{Model} & Acc\% $\uparrow$ & Sim $\uparrow$ & Acc\% $\uparrow$ & Sim $\uparrow$ & Acc\% $\uparrow$ & Sim $\uparrow$ \\
\midrule
\multicolumn{7}{l}{\textit{Search-based}} \\
NMR-Solver
& 6.27{\scriptsize $\pm$.00} & .218{\scriptsize $\pm$.000} & 13.96{\scriptsize $\pm$.00} & .348{\scriptsize $\pm$.000} & 17.37{\scriptsize $\pm$.00} & .401{\scriptsize $\pm$.000} \\
{\quad\quad\itshape +Formula Condition} 
& \underline{17.92{\scriptsize $\pm$.00}} & .276{\scriptsize $\pm$.000} & \underline{33.97{\scriptsize $\pm$.00}} & .411{\scriptsize $\pm$.000} & \underline{36.48{\scriptsize $\pm$.00}} & .428{\scriptsize $\pm$.000} \\
\midrule

\multicolumn{7}{l}{\textit{Transformer-based}} \\
CLAMS 
& 0.00{\scriptsize $\pm$.00} & .079{\scriptsize $\pm$.001} & 0.00{\scriptsize $\pm$.00} & .128{\scriptsize $\pm$.001} & 0.00{\scriptsize $\pm$.00} & .147{\scriptsize $\pm$.001} \\
NMRFormer 
& 1.75{\scriptsize $\pm$.04} & .225{\scriptsize $\pm$.000} & 2.81{\scriptsize $\pm$.05} & .296{\scriptsize $\pm$.000} & 3.30{\scriptsize $\pm$.03} & .323{\scriptsize $\pm$.000} \\
NMR2Struct 
& 0.24{\scriptsize $\pm$.08} & .220{\scriptsize $\pm$.004} & 1.05{\scriptsize $\pm$.19} & .337{\scriptsize $\pm$.006} & 1.85{\scriptsize $\pm$.27} & .387{\scriptsize $\pm$.006} \\
NMRMind 
& 11.75{\scriptsize $\pm$.17} & .370{\scriptsize $\pm$.000} & 23.22{\scriptsize $\pm$.05} & .547{\scriptsize $\pm$.000} & 27.00{\scriptsize $\pm$.15} & .591{\scriptsize $\pm$.000} \\
{\quad\quad\itshape +Formula Condition} 
& 15.49{\scriptsize $\pm$.08} & \underline{.408{\scriptsize $\pm$.001}} & 29.82{\scriptsize $\pm$.09} & \underline{.600{\scriptsize $\pm$.000}} & 34.03{\scriptsize $\pm$.09} & \underline{.644{\scriptsize $\pm$.001}} \\
\midrule

\multicolumn{7}{l}{\textit{Diffusion-based}} \\
DiffNMR 
& 0.00{\scriptsize $\pm$.00} & .174{\scriptsize $\pm$.000} & 0.00{\scriptsize $\pm$.00} & .254{\scriptsize $\pm$.000} & 0.00{\scriptsize $\pm$.00} & .285{\scriptsize $\pm$.000} \\
ChefNMR-S 
& 0.02{\scriptsize $\pm$.00} & .032{\scriptsize $\pm$.000} & 0.04{\scriptsize $\pm$.01} & .089{\scriptsize $\pm$.000} & 0.05{\scriptsize $\pm$.01} & .114{\scriptsize $\pm$.000} \\
ChefNMR-L(Finetune) 
& 1.93{\scriptsize $\pm$.02} & .136{\scriptsize $\pm$.001} & 4.36{\scriptsize $\pm$.08} & .259{\scriptsize $\pm$.000} & 5.66{\scriptsize $\pm$.01} & .301{\scriptsize $\pm$.000} \\
\midrule

\textbf{NMRAgent (Ours)} 
& \textbf{64.42{\scriptsize $\pm$.00}} & \textbf{.778{\scriptsize $\pm$.001}} & \textbf{66.29{\scriptsize $\pm$.00}} & \textbf{.788{\scriptsize $\pm$.001}} & \textbf{67.13{\scriptsize $\pm$.01}} & \textbf{.792{\scriptsize $\pm$.002}} \\
\bottomrule
\bottomrule
\end{tabular}
}
\end{table}

% \vspace{-6mm}

% \begin{table}[h]
% \caption{\textbf{Ablation study of NMRAgent modules.}}
% \vspace{-3mm}
% \centering
% \label{tab:ablation_study}
% \setlength{\tabcolsep}{8pt} % 调节列间距
% \begin{tabular}{ccccc}
% \toprule
% \toprule
% \textbf{R} & \textbf{D} & \textbf{F} & Acc\% $\uparrow$ & Sim $\uparrow$  \\
% \midrule
%  & & & 36.89 {\scriptsize $\pm$ .00} & NA {\scriptsize $\pm$ .00} \\
%  \checkmark & & & 60.47 {\scriptsize $\pm$ .00} & .758 {\scriptsize $\pm$ .000} \\
%  \checkmark & \checkmark & & 64.27 {\scriptsize $\pm$ .01} & .777 {\scriptsize $\pm$ .001} \\
% \checkmark & \checkmark & \checkmark & \textbf{64.42} {\scriptsize $\pm$ .00} & \textbf{.778} {\scriptsize $\pm$ .001} \\
% \bottomrule
% \bottomrule
% \end{tabular}
% \end{table}

\begin{table}[h]
\centering
\small
\caption{Top-$k$ accuracy comparison across heavy-atom-count buckets.}
\vspace{-3mm}
\label{tab:heavy_atom_bucket_appendix}
\setlength{\tabcolsep}{5pt}
\begin{tabular}{llrrrr}
\toprule
Method & Heavy atoms & Samples & Top-1 & Top-5 & Top-10 \\
\midrule
\multirow{4}{*}{NMRSolver}
& $<15$  & 521    & 19.00 & 28.98 & 30.13 \\
& 15--25 & 6,779  & 13.00 & 26.38 & 28.51 \\
& 25--35 & 9,607  & 12.63 & 29.20 & 32.13 \\
& $>35$  & 10,147 & 11.48 & 32.75 & 37.68 \\
\midrule
\multirow{4}{*}{NMRAgent}
& $<15$  & 521    & 14.40 & 33.21 & 39.54 \\
& 15--25 & 6,783  & 14.88 & \textbf{38.83} & \textbf{47.37} \\
& 25--35 & 9,617  & 15.35 & \textbf{42.66} & \textbf{52.60} \\
& $>35$  & 10,171 & \textbf{13.89} & \textbf{44.87} & \textbf{56.91} \\
\midrule
\multirow{4}{*}{NMRMind}
& $<15$  & 549    & \textbf{23.86} & \textbf{46.63} & \textbf{52.64} \\
& 15--25 & 6,748  & \textbf{19.12} & 36.04 & 41.70 \\
& 25--35 & 10,595 & \textbf{15.55} & 30.35 & 34.36 \\
& $>35$  & 9,192  & 12.51 & 23.80 & 27.18 \\
\bottomrule
\end{tabular}
\end{table}

% \vspace{-6mm}

\begin{table}[h]
\centering
\small
\caption{Top-$k$ accuracy comparison across benchmark datasets.}
\vspace{-3mm}
\label{tab:dataset_topk_appendix}
\setlength{\tabcolsep}{5pt}
\begin{tabular}{lrrrrrrrrr}
\toprule
Method 
& \multicolumn{3}{c}{NMRGym}
& \multicolumn{3}{c}{nmrshiftdb}
& \multicolumn{3}{c}{Exp450} \\
\cmidrule(lr){2-4}
\cmidrule(lr){5-7}
\cmidrule(lr){8-10}
& Top-1 & Top-5 & Top-10
& Top-1 & Top-5 & Top-10
& Top-1 & Top-5 & Top-10 \\
\midrule
NMRMind
& 10.10 & 15.70 & 18.30
& 1.90 & 4.10 & 5.40
& 1.77 & 2.88 & 2.88 \\

NMRSolver
& 17.92 & 33.97 & 36.48
& 20.40 & 33.10 & 37.90
& 52.40 & 65.10 & 66.90 \\

NMRAgent
& \textbf{64.40} & \textbf{66.29} & \textbf{67.13}
& \textbf{67.56} & \textbf{78.10} & \textbf{81.05}
& \textbf{61.60} & \textbf{68.90} & \textbf{70.00} \\
\bottomrule
\end{tabular}
\end{table}

\end{document}